%% file: main.tex
\documentclass[10pt,conference]{IEEEtran}
\IEEEoverridecommandlockouts
\usepackage{cite}

\usepackage{mathtools}
\usepackage{graphicx}
\usepackage{subcaption}
\DeclarePairedDelimiter{\ceil}{\lceil}{\rceil}
\usepackage{amsmath,amssymb,amsfonts}
\usepackage{algorithmic}

\usepackage{textcomp}
\usepackage{xcolor}
\usepackage{comment}
\def\BibTeX{{\rm B\kern-.05em{\sc i\kern-.025em b}\kern-.08em
    T\kern-.1667em\lower.7ex\hbox{E}\kern-.125emX}}
\usepackage{breqn}

\newcommand\norm[1]{\left\lVert#1\right\rVert}

\begin{document}
\title{Active Learning Under Malicious Mislabeling and Poisoning Attacks}

\author{\IEEEauthorblockN{Jing Lin}
\IEEEauthorblockA{\textit{ICNS Lab and Cyber Florida}\\
\textit{University of South Florida}\\
Tampa, FL. USA \\
jinglin@usf.edu}
\and
\IEEEauthorblockN{Ryan Luley}
\IEEEauthorblockA{\textit{High Performance Systems Branch} \\
\textit{U.S. Air Force Research Laboratory}\\
Rome, New York, USA \\
Ryan.Luley@us.af.mil}
\and
\IEEEauthorblockN{Kaiqi Xiong}
\IEEEauthorblockA{\textit{ICNS Lab and Cyber Florida}\\
\textit{University of South Florida}\\
Tampa, FL. USA \\
xiongk@usf.edu}

\thanks{Received and approved for public release by AFRL on 18 November 2020, case number AFRL-2020-0311.}
}
\maketitle
\begin{abstract}
\input{abstract.tex}
\end{abstract}

\section{Introduction}\label{sec:introduction}\input{introduction}
\section{Background and Related Work}\label{sec:related}\input{related}

\section{Threat Model and Assumptions}\label{sec:3}\input{3}
\section{Methodology}\label{sec:design}\input{design}
\section{Evaluation}\label{sec:evaluation}\input{evaluation}
\section{Conclusions and Future work} \label{sec:conclusion}\input{conclusion}
\section*{Acknowledgment}\label{sec:ack}\input{ack}

\bibliographystyle{IEEEtran}
\bibliography{reference}
\end{document}

%% file: abstract.tex
Deep neural networks usually require large labeled datasets for training to achieve state-of-the-art performance in many tasks, such as image classification and natural language processing. Although a lot of data is created each day by active Internet users, most of these data are unlabeled and are vulnerable to data poisoning attacks. In this paper, we develop an efficient active learning method that requires fewer labeled instances and incorporates the technique of adversarial retraining in which additional labeled artificial data are generated without increasing the budget of the labeling. The generated adversarial examples also provide a way to measure the vulnerability of the model. To check the performance of the proposed method under an adversarial setting, i.e., malicious mislabeling and data poisoning attacks, we perform an extensive evaluation on the reduced CIFAR-10 dataset, which contains only two classes: airplane and frog. Our experimental results demonstrate that the proposed active learning method is efficient for defending against malicious mislabeling and data poisoning attacks. Specifically, whereas the baseline active learning method based on the random sampling strategy performs poorly (about 50\%) under a malicious mislabeling attack, the proposed active learning method can achieve the desired accuracy of 89\% using only one-third of the dataset on average.

%% file: introduction.tex
The performance of a deep learning model depends on the size of a labeled dataset, model complexity, and so on. Though we generated about 2.5 quintillion bytes of data each day from a variety of cyber and physical systems~\cite{website}, a large portion of this data is unlabeled. To use it for supervised learning requires human labelers to take time and effort to provide the labels for each data point. Because the 2.5 quintillion bytes of data per day results in enormous unlabeled data, we need to find a way to effectively select a portion of data for annotating. Active learning was born from such needs.

Either experienced labelers (such as doctors for medical data) or ordinary labelers (such as graduate students for image data) can hand-annotate the data depending on a classification task. Despite the level of expertise, all labelers can make a mistake while labeling. Fard, et al.~\cite{fard2017impact} has shown that deep learning networks are robust against minor uniform mislabeling, randomly mislabeling one class as any other class. However, systematic mislabeling, consistently mislabeling a class as another class, is more far-reaching~\cite{fard2017impact}. In this paper, we consider both human error and malicious mislabeling in active learning. 

In addition to the mislabeling, another security concern is a data poisoning attack~\cite{mukherjee2020guarantees}. Despite state-of-the-art performance, deep learning models are easily fooled by data poisoning attacks and evasion attacks~\cite{yuan2019adversarial}. The main difference between data poisoning attacks and evasion attacks is the assumption of an attacker's ability. For a data poisoning attack, an adversary tries to manipulate some training data in order to change a decision boundary and cause the misclassification of a specified test instance on a targeted attack. Contrarily, an adversary tries to generate an adversarial example by introducing a subtle perturbation to a specified test instance in order to mislead a learning algorithm and evade detection in an evasion attack. In this paper, we propose a novel active learning method that mitigates the effect of data poisoning attacks by utilizing adversarial examples. 

The key contributions of this paper are in the following: 

\begin{itemize}
    \item We present a new heuristic methodology for robust active learning that utilizes the adversarial examples as a measure of uncertainty and a minimax algorithm for data selection. Furthermore, the crafted adversarial examples can be added to a training set without incurring an extra labeling budget.
    \item We develop a confidence score updating scheme to check the performance of each labeler. This scheme significantly reduces the effectiveness of a malicious mislabeling attack. Specifically, the proposed active learning method achieves the desired accuracy of 89\% using only one-third of the labeled data, whereas the baseline active learning method performs poorly (50\%).
    \item We empirically demonstrate that the proposed methodology is highly efficient for defending against malicious labeling and targeted poisoning attacks.  
\end{itemize}

The remainder of this paper is organized as follows. In section \ref{sec:related}, we discuss existing active learning methods. We describe a threat model and assumptions in section \ref{sec:3}. In section \ref{sec:design}, we describe in detail the proposed approach, followed by its evaluation in section \ref{sec:evaluation}. Section \ref{sec:conclusion} 
discusses our conclusion and provides future work. 

%% file: related.tex
This section provides a brief description of existing active learning methods.
Though gathering large quantities of unlabeled data is easy and cheap, obtaining the label for them is usually time-consuming and expensive. Active learning reduces the number of labeled instances needed for training a model effectively. Active learning is an iterative process that consists of training and querying. First, a model is trained based on an initially labeled training set. After the initial model is trained, a query strategy is used to select a new subset of data for label querying. Then, an oracle, e.g., a human annotator, labels these data, and the model is retrained with currently available labeled data. The process repeats until a stopping criterion is met, such as the reach of the desired accuracy. A variety of active learning algorithms have developed over the years. They can be broadly categorized as uncertainty-based active learning and diversity-based active learning.

The uncertainty-based active learning algorithms focus on selecting instances based on an uncertainty principle \cite{nguyen2004active}, \cite{settles2009active}, \cite{huijser2017active}. The uncertainty principle states that a training instance that is most uncertain to a model contains the most new information for improving the model. There are many ways of measuring uncertainty. For instance, the distance between an instance and the classification boundary can be used to measure the uncertainty. As an instance is closer to the classification boundary, it is more difficult for a classifier to make a high confidence classification. This provides a measure of uncertainty.  Another measure of uncertainty is the final confidence provided by a trained model, as the lower confidence provided by a model indicates a relatively high uncertainty in its performance. An alternative way is to measure the difference between the two highest confidence levels provided by the model. The idea behind this way is that if the model is confident in its prediction, it should only provide a very high confidence value for one class and low confidence values for other classes. On the contrary, if a model is unsure of its prediction, it generally provides relatively high confidence values for two or more classes.  Other measures of the uncertainty include entropy \cite{joshi2009multi}, smallest predicted class probability \cite{wang2014new}, and so on; See, e.g., \cite{settles2009active} for an overview of classic active learning methods. 

However, recent work on data poisoning attacks, e.g., \cite{ducoffe2018adversarial}, \cite{truong2020systematic}, \cite{zhu2019transferable}, has shown that uncertainty sampling may help attackers in achieving their goals. Since poisoning instances are usually near the decision boundary, they are more likely to be selected by the uncertainty sampling approaches.

Another class of active learning algorithms focuses on selecting the most diverse and representative dataset. For instance, Yang et al. \cite{article} proposed an optimization-based active learning algorithm that imposed a diversity constraint objective function. The core-set approach \cite{sener2017active} is a diversity-based approach using the core-set selection. Batch Active learning by the Diverse Gradient Embeddings (BADGE) algorithm \cite{ash2020deep} selects a subset of instances whose gradients with respect to the parameters in the output layer are most diverse. In this research, we present a new active learning method that incorporates both the representative sampling and uncertainty sampling ideas.

%% file: 3.tex
In this paper, we study the effect of a mislabeling attack and a data poisoning attack on the active learning of deep neural networks. There is a large amount of data generated every day. However, these data need to be properly labeled before we can apply supervised machine learning algorithms. This is not always feasible with a restricted labeling budget. Crowdsourcing is not reliable as humans make mistakes as well, and the adversary can intentionally provide misleading labels. Furthermore, the data obtained could be poisoned. 

In this work, we assume that the class labels provided by a human labeler are not perfect. That is, we consider the effect of human errors. We assume each benign labeler mislabels an instance $p_r \in (0, 1)$ of the time. Moreover, it is equally likely that an instance (whose true class is $c$) is mislabeled as any other class $\bar{c} \in C/{c}$, where $C$ is the set of classes. Each independent labeler will provide the true label with probability $1-p_r$ and the remaining probability $p_r$ is evenly distributed to other class labels. That is, if an instance $x$ belongs to class $c$, then a human labeler will label it 
\[
  O(x) =
  \begin{cases}
  \bar{c} & \text{if $u<p_r$} \\
c & \text{if $u \geq p_r$} ,
  \end{cases}
\]
where $u$ is a uniform random number between 0 and 1, and $\bar{c}$ is randomly selected from the set $C/{c}$.
However, under a malicious mislabeling attack, an attacker consistently mislabels instances of a target class $c_t$ as if it belongs to a benign class $\bar{c}_t\in C/c_t$. As shown in \cite{fard2017impact}, biased labeling usually severely reduces the classification accuracy of the target class but has little effect on the non-target classes.

Furthermore, we consider a data poisoning attack. The attacker's objective is to inject a data poisoning instance $p$ such that
\begin{equation}\label{eq:0}
  p = \operatorname*{argmax}_{p\in D_i} \mbox{\textbf{C}}(F(\theta_i^*),  D_{t}),
\end{equation}
where \mbox{\textbf{C}} is a cost function that measures the performance of a model $F$ on a target set $D_t$, and the parameters $\theta_i^*$ of the model $F$ at iteration $i$ are determined by minimizing a loss function $L$ used to train a poisoned training set $D_i$ as follows:
\begin{equation}
\theta_i^* = \operatorname*{argmin}_{\theta_i} L(F(\theta_i, D_i)).
\end{equation}

The poison instance $p$ can be generated by solving Equation \ref{eq:0} based on \cite{2018arXiv180400792S}. The targeted poisoning attack (also known as the feature collision attack) proposed in \cite{2018arXiv180400792S} assumes an attacker has no knowledge of the training data but has knowledge of the model architecture and its parameters. In this research, we extend this poisoning attack to an active learning setting, where the attacker can inject poison instances into the data pool. Specifically, the attacker can generate a poison instance $p$ by solving the following optimization problem:
\begin{equation}\label{eq:1}
    p= \operatorname*{argmin}_x \norm{ f_i(x) -f_i(t) }^2_2+\beta \norm{x -b }^2_2,
\end{equation}
where $f_i$ is the output of the penultimate layer of a network obtained at iteration $i$ of the active learning process, $\beta \in [0,1]$ is a similarity parameter that indicates the relative importance of the first component $\norm{ f_i(x) -f_i(t) }^2_2$ to the second component $\norm{x -b }^2_2$, $t$ is a target instance, and $b$ is a base instance. The research problem of this study is to propose a novel defense method against such an attack under the assumption that the oracle does not always provide correct/true labels.

%% file: design.tex
\begin{figure}[hp]
\centering
\includegraphics[width=0.48\textwidth]{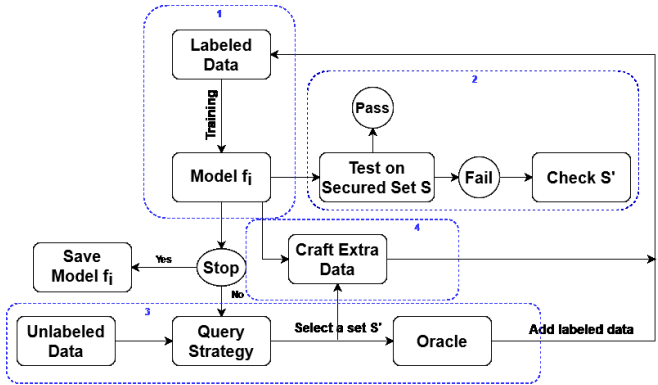} 
\caption{The Proposed Active Learning Framework }
\label{2020}
\end{figure}

In this section, we propose a robust active learning method that reduces the effectiveness of malicious mislabeling and data poisoning attacks. Fig.~\ref{2020} shows the framework of our proposed active learning method. The four main components of this framework are (1) training, (2) performance checking, (3) querying, and (4) artificial data crafting. The training step is the same as regular training on a labeled dataset. Initially, a set of $m$ data points $D$ is randomly selected to query for labels. Then, the model is trained using the labeled dataset $D$. To check the performance of the model and the existence of malicious instances, a representative subset $S$ of data with labels is set aside and secured so that adversary does not know it. This set $S$ is updated every $q$ iterations to make sure it represents the diversity of the current dataset, where $q$ is an application-specific parameter that depends on the dynamic of the dataset. If the underlying distribution of the data does not change over time and there is no problem with its security, $q=\infty$ and no update is needed.

Besides, for each queried instance, we can generate both adversarial examples and data augmentation. In this way, the training set is increased without incurring an extra query budget. The various adversarial crafting techniques and data argumentation techniques are useful in this case. For each center $c$, $N_c$ instances are selected. There is a total of $\sum_{c=1}^n N_c$ data points $S^\prime$ from $n$ centers. Then, a group of human labelers provides the labels for these data points. Next, this newly labeled dataset 
\begin{equation}
    D'=\{(x, O(x))|x \in S^\prime\},
\end{equation} 
where $O(x)$ is the label provided by an oracle/human labeler for an instance $x$, and $D'$ is added to the previously obtained labeled dataset $D$; i.e.,
$D=D \cup D^\prime.$
Next, we train the model $F_i$ on the new dataset $D$. The performance of the updated model is tested on a secured set $S.$ If the performance degrades by a threshold $\epsilon$, the newly obtained dataset $D'$ is examined by $h$ ($h>0$) independent labelers to see if any instance is poisoned or mislabeled.
The probability $P_h$ that at least one labeler out of $h$ ($h>0$) independent labelers provides correct annotation is  
\begin{equation}\label{eq:3}
    P_h=1-(1-P_{1})^h,
\end{equation} 
where $P_{1}$ is the average performance of a human labeler.
To find the number $h$ of independent labelers to ensure that $P_h$ is large enough, saying $0.9995$, we solve equation \ref{eq:3} and obtain 
 $$h = \ceil[\Bigg] {\frac{ \log{(1-P_h)} } { \log{ (1- P_1} )} }.$$
For instance, \cite{ho2018cifar10} showed the human annotation performance for the CIFAR-10 dataset is 93.91\%. Hence, three independent labelers are enough to ensure that at least one labeler provides correct annotation. If the performance does not degrade more than $\epsilon$, the model is accepted and stopping criterion is checked. If the stopping criterion is met, the process stops.  Otherwise, the process repeats.
The stop criterion can be (1) the minimal perturbation of all adversarial examples is larger than a threshold $\tau$, (2) the desired classification accuracy is reached, or (3) the labeling budget is exhausted.

In the following subsections, we provide additional discussions for performance checking, querying, and artificial data crafting.

\subsection{Performance Checking}

The ultimate goal of malicious labeling attacks is to reduce the accuracy of a model. Hence, the best way to detect it is by monitoring the classification accuracy on a secured subset $S$. If the performance (i.e., classification accuracy) of a newly trained model $F_i$ on $S$ drops by a threshold $\epsilon$ compared to $F_{i-1}$, then the newly queried dataset used to train $F_i$ is likely malicious. They need to be further examined. Otherwise, accept $F_i$.  

For further examination, three independent labelers are assigned to relabel the newly queried data point(s). For each data point, if there is any inconsistency in the label, the data point is discarded. However, if the consistent label is provided and it is different from the original label provided by the original labeler $O$, then the confidence in $O$'s label is reduced. The degree of reduction in confidence depends on whether $F_i$ provides high confidence or low-confidence misclassification. If $F_i$ provides a low-confidence value for this misclassification, then it may be a mild drift and we should reduce the confidence in $O$'s label by $a$ amount, where $a$ is inversely proportional to the confidence value provided by $F_i.$ If $F_i$ provides a high-confidence value for this misclassification, then we should be alert and the trustworthiness of $O$ needs to be checked.

\subsection{Querying Strategy}

The proposed querying strategy looks for unlabeled instances that are different from the most labeled ones. For this purpose, a clustering algorithm, such as k-mean, k-medoids (PAM), and hierarchical clustering, can be used to find $n$ centers, where $n$ can be determined by various methods such as elbow, silhouette, or gap statistic methods. For each center $c$, $N_c$ unlabeled instances whose corresponding distance (in a feature space) from the currently labeled instances is the largest are selected; i.e., select $x \in A$ by solving the following optimization problem:
\begin{equation}\label{eq:2}
    \max_{x_1 \in A_c} \min_{x_2 \in D} d(x_1,x_2),
\end{equation}
where $A_c$ is the collection of unlabeled instances that belong to cluster c, and $d$ is a distance metric in the feature space. This ensures the diverse instances being selected.

\subsection{Artificial Data Crafting}

To take advantage of the ``indistinguishable property" of adversarial examples, we utilize various adversarial crafting techniques to generate extra data for training. Using the labeled instances, we can generate additional instances that are similar to these labeled instances without incurring an extra labeling budget. Since the minimal adversarial perturbation obtained through adversarial crafting techniques can misclassify these instances, they point at the vulnerable area of the classifier and are a great place to get more data points for the next iteration of active learning. 

%% file: evaluation.tex
We empirically evaluate the proposed defense method on a subset of the CIFAR-10 dataset~\cite{krizhevsky2009learning}. The baseline model for comparison is active learning based on random selection. Our experiments were conducted in the high-performance computing system within the campus private cloud at the University of South Florida. In the set of experiments, we wish to address the following: (1) Does our proposed method perform as well as the standard active learning method when there is no attack? We try to address this question since a few papers (e.g., \cite{akhtar2018threat}, \cite{pang2020tale}) have shown the trade-off between the performance of the active learning algorithms with attacks and their performance without attacks. (2) Is our proposed method effective under a malicious mislabeling attack? and (3) Is our proposed method effective under a poison attack? That is, are certain vulnerabilities (e.g., to specific inputs) learned when we perform the adversarial active retraining? 

\subsection{Dataset and Model}

The CIFAR-10 dataset~\cite{krizhevsky2009learning} is an image classification dataset that consists of ten classes: airplane, automobile, bird, cat, deer, dog, frog, horse, ship, and truck. Because of the limited computation resource and the slowness of the data poisoning attack method proposed in \cite{2018arXiv180400792S}, we reduce the CIFAR-10 dataset to the CIFAR-2 dataset, only considering two classes, airplane and frog. To further speed up the data poisoning attack, we reduce the training set to 3,000 training instances instead of 10,000. As you can see in the evaluation section, the proposed method does not even need to have 3,000 training instances to reach the desired accuracy. Thus, this limitation speeds up the evaluation time without affecting the performance. The testing set has 2,000 instances, and each testing instance is considered as a targeted instance during the poison instance generation process.

The network architecture for the CIFAR-10 dataset is based on ResNet50 \cite{he2016deep}. More precisely, ResNet50 is used for feature extraction, and then a dropout rate of 0.2 is applied to it. Last, a fully connected layer with the sigmoid activation function is used for binary classification. 

\subsection{Artificial Data Generation and Data Poisoning Attack }

Adversarial examples and poison instances are generated using the Adversarial Robustness Toolbox (ART v1.3) \cite{art2018}. C\&W, PGD, and DeepFool attacks are used to generate artificial images in addition to Gaussian noise. The PGD attack is selected over others because the PGD attack generates the strongest first order attack \cite{madry2017towards}. However, due to the fact that the PGD attack implemented in ART does not take advantage of GPU power, the runtime for the PGD attack is much higher than the C\&W attack. Therefore, the PGD attack is not included during the training. The maximum number of iterations for the C\&W attack and DeepFool are 20 and 100 (default in ART), respectively. Random noise added to a training image follows a Gaussian distribution with mean 0 and variance 0.3. This is a reasonable limit on what is permissible because a larger perturbation would distort the image too much.

We generate the data poisoning attack based on the scheme in~\cite{2018arXiv180400792S} to evaluate the robustness of the proposed active learning method. This is the only data poisoning attack on a DNN model available in ART. We consider each test instance as a targeted instance. For each targeted poisoning attack, we use first 50 training instances from a different class than the targeted instance as base instances for generating the poison instances. A small amount of watermark (0.1-0.3) is added to increase the attack success rate while the poison instances are indistinguishable from their corresponding base instances. The exact amount of the watermarks added depends on the difficulty of each targeted attack scenario. After a set of poison instances is generated for each targeted attack, we check whether the attack is successful. If a targeted poisoning attack is not successful, the corresponding poison instances are discarded. In short, we only save the set of poisoning instances that mislead the baseline model. In the end, we generate 67 successful data poisoning images, which can be used to evaluate the performance of the proposed active learning method. 

\subsection{Mislabeling} \label{mislabel}

For most deep learning tasks, we assume that the labels provided are correct. However, in the real world, humans make mistakes and adversaries can inject malicious labeling~\cite{taheri2020defending, paudice2018label}. In this paper, we consider both cases. First, we assume a human labeler has a 5\% chance to make a mistake when annotating. The 5\% is used since the dataset used is  low-resolution images that are difficult to be classified correctly sometimes. Furthermore, with a total of eleven labelers for the experiments, one of them is malicious. More precisely, a malicious labeler will intentionally misclassify an image in the `airplane' class as the `frog' class.

\subsection{Results}

In this section, we evaluate the performance of the proposed active learning method by addressing the three questions posed at the beginning of the evaluation section.

\subsubsection{Does our proposed active learning method perform as well as the standard active learning method when not under an attack?}

\begin{table}[ht]
\caption{Statistics of Labeling Budgets}
\begin{center}
\begin{tabular}{|l|l|l|}
 \cline{1-3}
Model      &Baseline & Proposed \\ \cline{1-3}
Mean     &1186 &572.3 \\ \cline{1-3}
Standard Deviation&182.6&51.7 \\ \cline{1-3}
Minimum &608 &462 \\ \cline{1-3}
25\% & 1128&533 \\ \cline{1-3}
50\%      &1184 & 563 \\ \cline{1-3}
75\%      &1272 & 610 \\ \cline{1-3}
Maximum      &1536 & 672 \\ \cline{1-3}
\end{tabular}
\end{center}
\label{t1}
\end{table}
Here, we consider only the benign mislabeling caused by human errors.
The prediction accuracy of the classifier trained with the full training set is 89\%. Using it as the desired accuracy, we find the labeling budget required to achieve such accuracy. We compare our proposed active learning method with the active learning based on a random sampling strategy (baseline). The experiment is repeated 30 times, and the summary statistics are provided in Table~\ref{t1}. As shown in the table, the proposed active learning method only needs about 48\% of the labeling budget required by the baseline active learning method when there are no attacks. Furthermore, the variation in performance is much lower for our proposed method compared to the baseline model. The maximum number of labeling budgets required to achieve the desired accuracy is halved using our proposed method.

\subsubsection{Is our defense method effective under a malicious mislabeling attack?}
\begin{figure*}
\subfloat[]{\includegraphics[width = 2.38in]{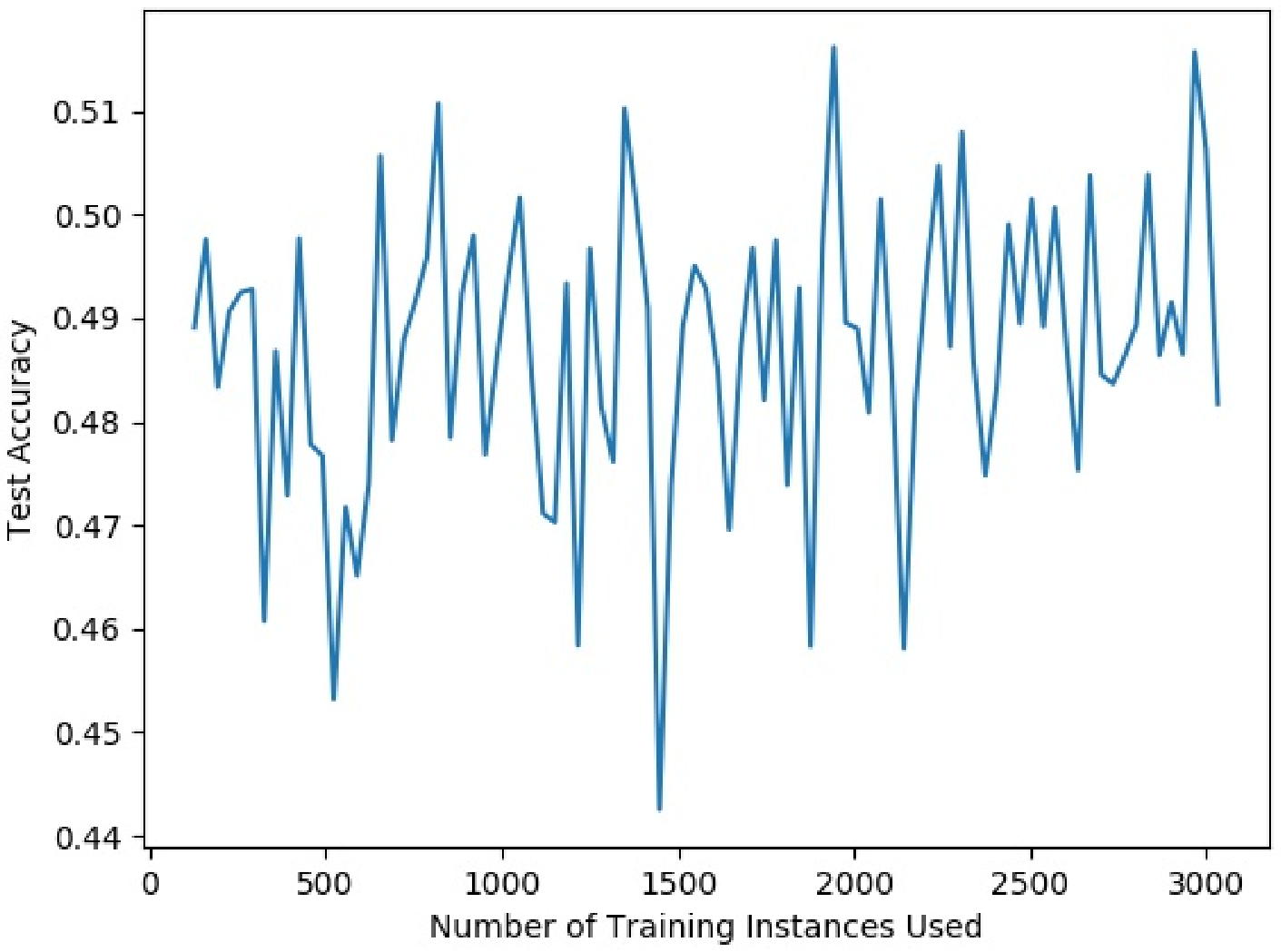}} 
\subfloat[]{\includegraphics[width = 2.38in]{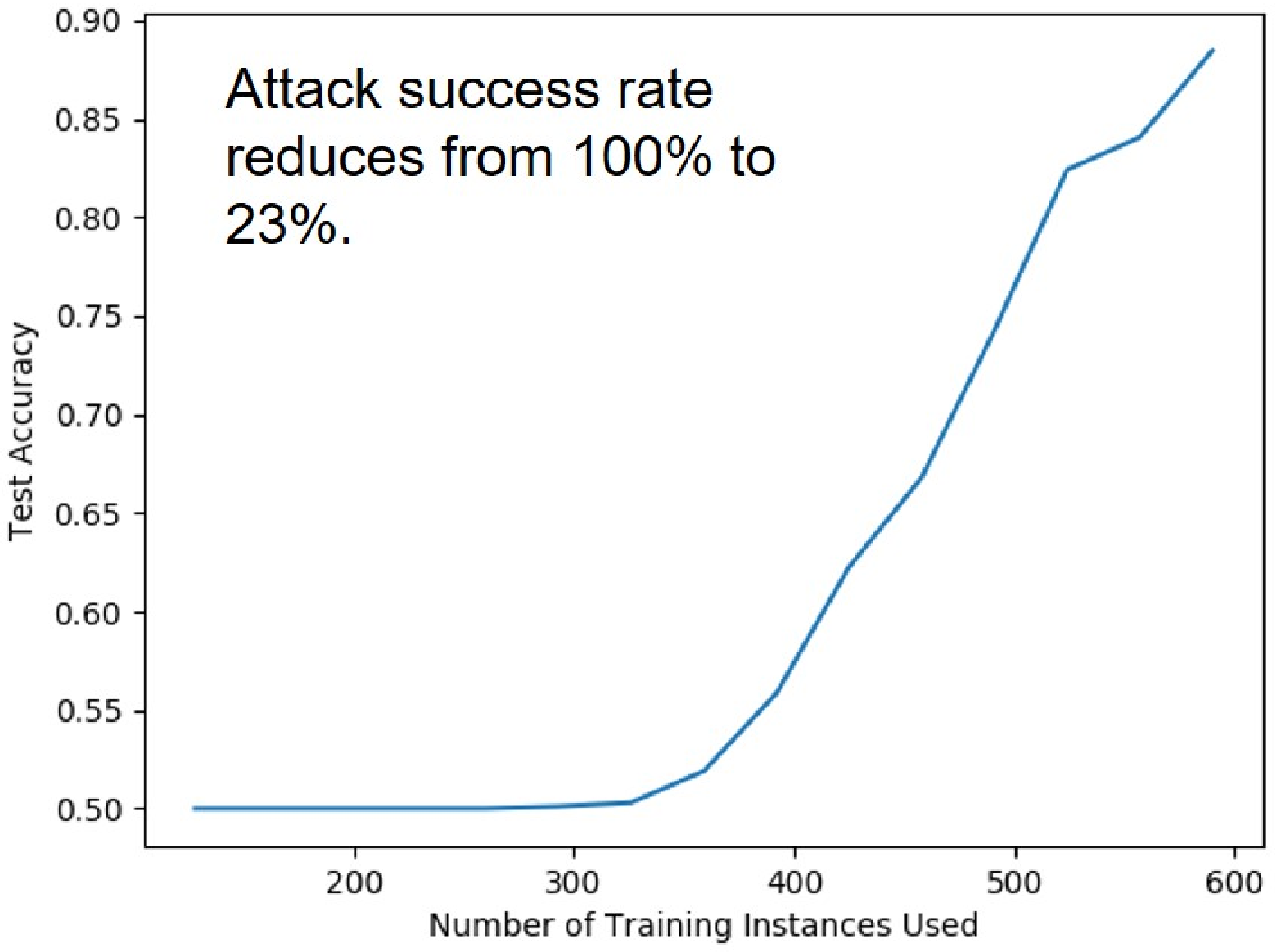}}
\subfloat[]{\includegraphics[width = 2.38in, height=4.5cm]{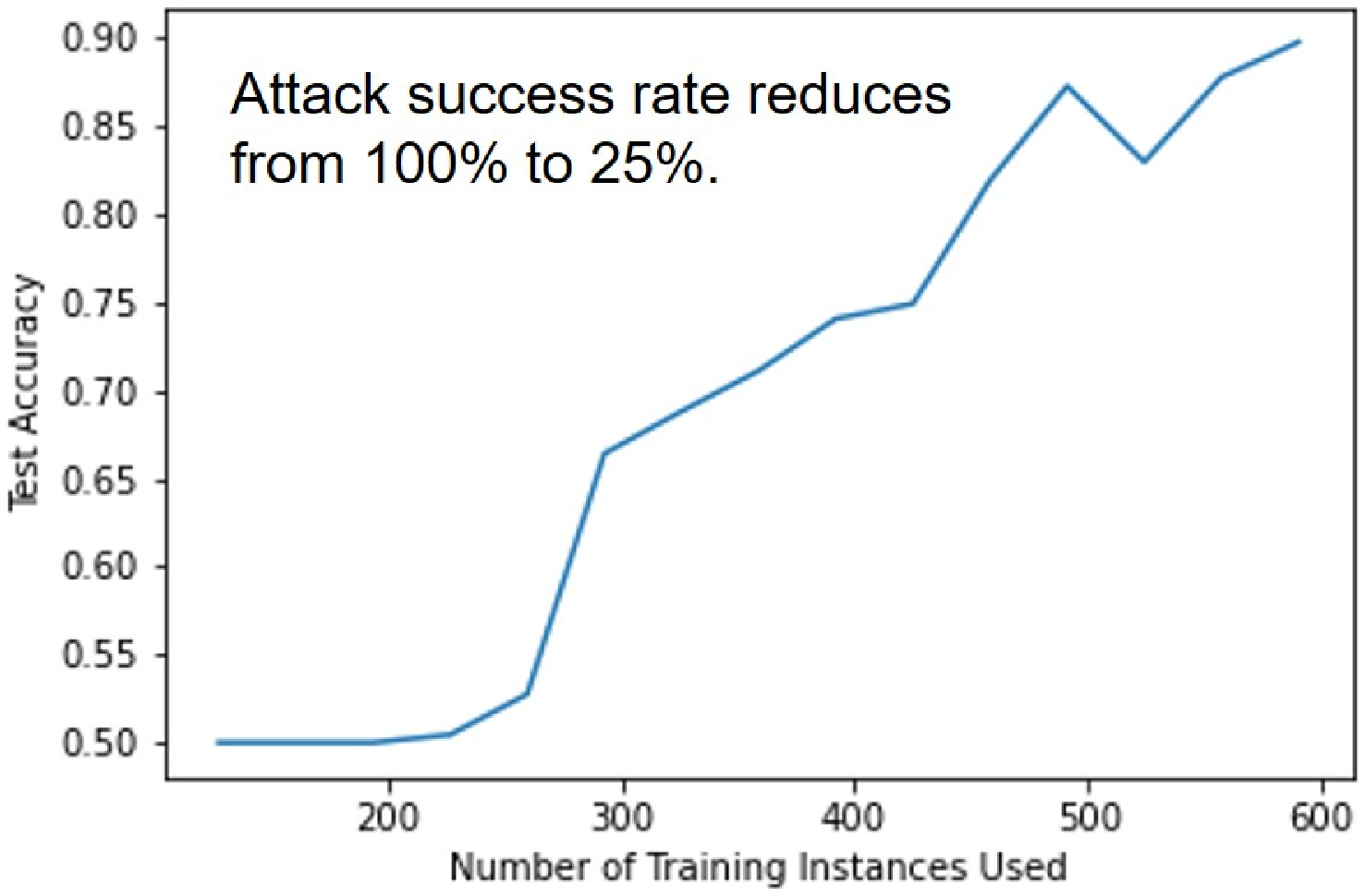}} 
\caption{(a) Baseline model performance under a malicious labeling attack. (b) Baseline model performance under a poisoning attack. (c) Robust active learning model performance under a poisoning attack }
\label{5}
\end{figure*}

First, we evaluate the performance of the baseline model under a malicious mislabeling attack (see subsection \ref{mislabel} for the implementation of a malicious mislabeling attack). Similar to the previous experiment, we repeated our experiments 30 times. However, the baseline model is not able to achieve the desired accuracy in all the experiments. As shown in Fig. \ref{5}(a), the malicious mislabeling attack is very effective against the baseline model. The test accuracy fluctuates between 0.44 and 0.52. Even when we use all 3,000 training instances, the test accuracy is still close to 0.5. Nevertheless, our proposed active learning method is robust against the malicious mislabeling attack. We only need about one-third of the original training set size to achieve the same accuracy (89\%) even under a  malicious mislabeling attack.

\subsubsection{Is our proposed method effective under a poison attack?}

Here, we consider a targeted poison attack and evaluate the performance of the baseline and our proposed active learning method under such an attack. The attacker tries to reduce the performance of the classifier on the targeted instance while maintaining the performance of the classifier on the clean data. This makes it very difficult to detect as the classification accuracy on the clean data is not changed under the attack. However, the classifier makes misclassification on the targeted instance. 

Since most uncertainty-based active learning strategies select unlabeled instances that are closer to the decision boundary, this makes these uncertainty-based active learning strategies more vulnerable to the data poisoning attack. On the contrary, simple active learning based on a random sampling strategy is more robust against such an attack. Hence, we will compare our proposed active learning method with active learning based on a random sampling strategy. The result indicates the test accuracies of the two active learning strategies under the poisoning attack are similar though the labeling budget required is halved if our proposed method is used (Fig. \ref{5} (b)-(c)). Furthermore, the proposed method has a much lower variation in labeling budgets than the baseline method. 

%% file: conclusion.tex
Recent work has shown that machine learning models are vulnerable to adversarial examples and data poisoning attacks. In this work, we utilized these adversarial examples to generate artificial data for training without incurring an extra labeling budget. Furthermore, we proposed a confidence score updating system to check the trustworthiness of each labeler and a performance check on a secured set to monitor the potential mislabeling attack. The experiment results showed that our proposed active learning method is robust against malicious mislabeling and data poisoning attacks. Though we only evaluated the performance of the proposed active learning method on binary classification in the campus private cloud, the method can be applied to a multi-classification task in a distributed system of a public cloud as well. Although most classic single-query active learning strategies do not work as well as a random sampling strategy on deep neural network and many uncertainty-based active learning strategies also perform poorly under data poisoning attacks, it would be interesting to compare the proposed method with the state-of-the-art batch active learning method for CNNs, core-set \cite{sener2017active} in the future. Furthermore, our evaluation is based on a single dataset. It would be interesting to check the effectiveness of the method on other datasets. Last, the artificial crafting step introduces additional computation cost. Hence, future work will also investigate the scalability of the proposed method. We can parallelize the process and utilize the scalability in cloud computing for this purpose. Besides a private cloud, we can also deploy our proposed method on the public cloud to not only achieve scalability but also to increase reliability and flexibility.

%% file: ack.tex
We acknowledge the AFRL Internship Program to support Jing Lin\textquotesingle s work and the National Science Foundation (NSF) to partially sponsor Dr. Kaiqi Xiong\textquotesingle s work under grants CNS 1620862 and CNS 1620871, and BBN/GPO project 1936 through an NSF/CNS grant. This material is based on research sponsored by the Air Force Research Laboratory under agreement number FA8750-20-3-1004. The U.S. Government is authorized to reproduce and distribute reprints for Governmental purposes notwithstanding any copyright notation thereon. The views and conclusions contained herein are those of the authors and should not be interpreted as necessarily representing the official policies or endorsements, either expressed or implied, of the Air Force Research Laboratory or NSF.